\documentclass{article}
\usepackage{spconf,amsmath,epsfig}
\usepackage{comment}
\usepackage{color}
\usepackage{adjustbox}
\let\OLDthebibliography\thebibliography
\renewcommand\thebibliography[1]{
  \OLDthebibliography{#1}
  \setlength{\parskip}{0pt}
  \setlength{\itemsep}{0pt plus 0.3ex}
}

\begin{document}\sloppy

\def\x{{\mathbf x}}
\def\L{{\cal L}}

\title{Knowledge Transfer Based Fine-Grained Visual Classification}
%
\name{Siqing Zhang, Ruoyi Du, Dongliang Chang, Zhanyu Ma (*), Jun Guo}
\address{}

\maketitle

\begin{abstract}

Fine-grained visual classification (FGVC) aims to distinguish the sub-classes of the same category and its essential solution is to mine the subtle and discriminative regions. Convolution neural networks (CNNs), which employ the cross entropy loss (CE-loss) as the loss function, show poor performance since the model can only learn the most discriminative part and ignore other meaningful regions. Some existing works try to solve this problem by mining more discriminative regions by some detection techniques or attention mechanisms. However, most of them will meet the background noise problem when trying to find more discriminative regions. In this paper, we address it in a knowledge transfer learning manner. Multiple models are trained one by one, and all previously trained models are regarded as teacher models to supervise the training of the current one. Specifically, a orthogonal loss (OR-loss) is proposed to encourage the network to find diverse and meaningful regions. In addition, the first model is trained with only CE-Loss. Finally, all models' outputs with complementary knowledge are combined together for the final prediction result. We demonstrate the superiority of the proposed method and obtain state-of-the-art (SOTA) performances on three popular FGVC datasets.


\end{abstract}
\section{Introduction}
\label{sec:intro}
Compared with general image classification tasks, fine-grained visual classification (FGVC) aims to identify sub-classes of a given object class \cite{fu2017look,zheng2017learning,yang2018learning,ge2019weakly,zhang2019learning}. Due to different sub-classes of a common visual class only differ in subtle details, FGVC faces the challenges of large intra-class variations and small inter-class variations \cite{fu2017look,zhang2019learning}. 

\begin{figure}[!t]
\centering
\includegraphics[height=3.5cm]{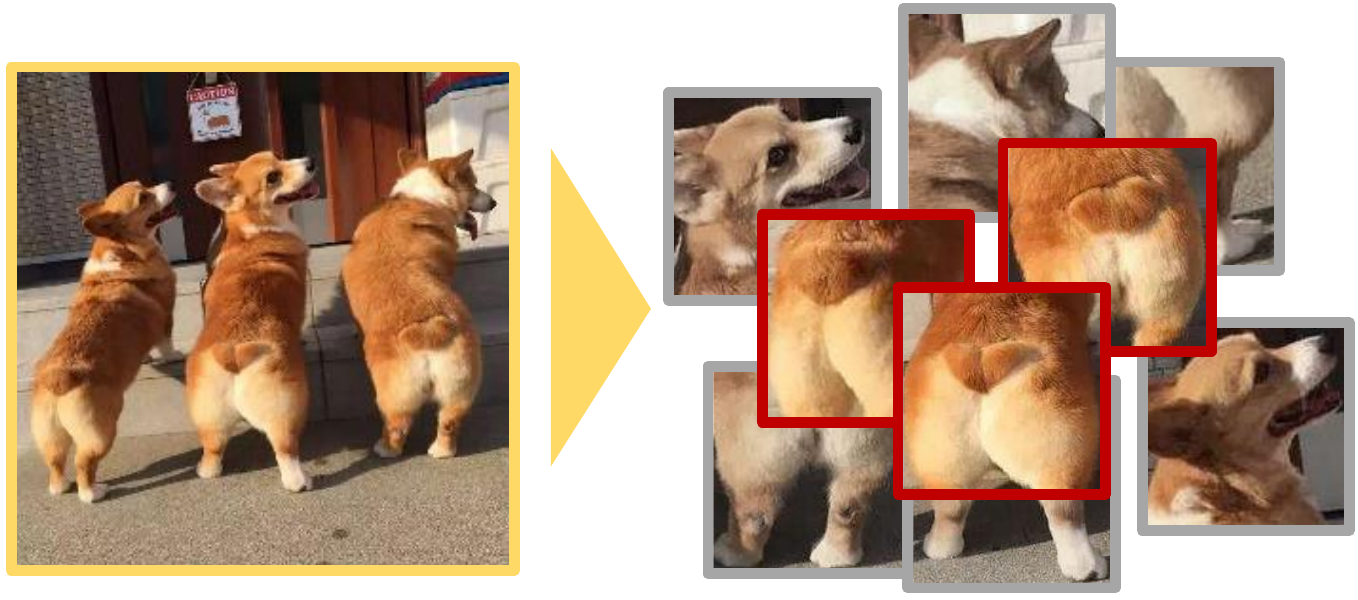}
\caption{For Corgis, round butts are their most significant features, but we can also easily recognize them through its cute heads or short legs.}
\label{fig:corgi}
\end{figure}

\begin{figure*}[!t]
\centering
\includegraphics[height=5cm]{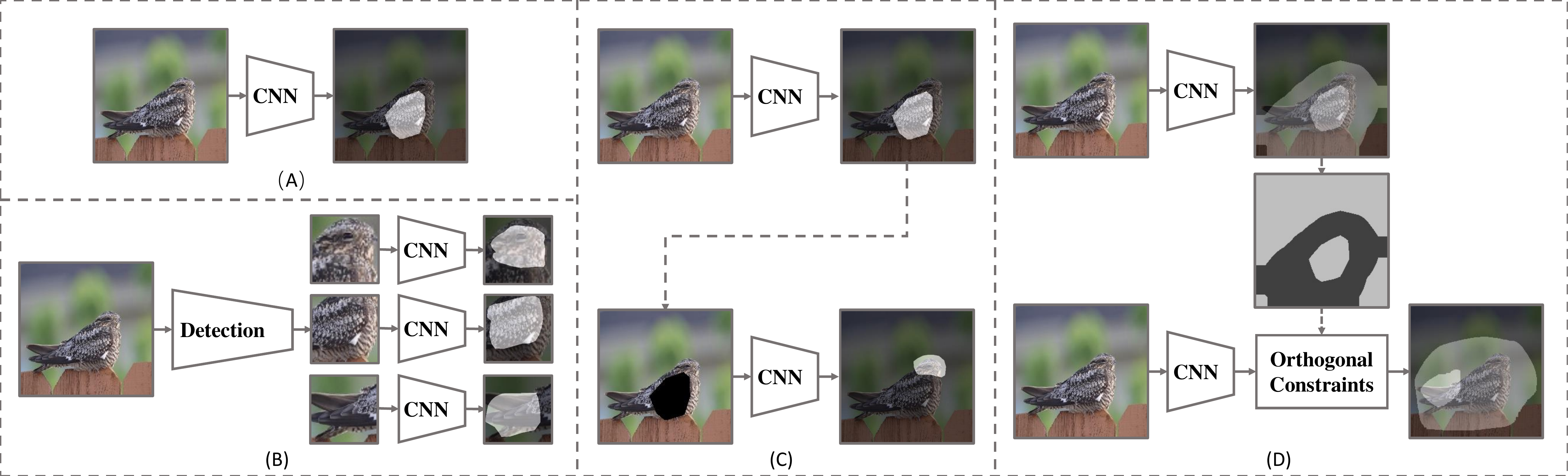}
\caption{Illustration of features learned by general methods (A B, and C) and our proposed method (D). (A) Convolution neural networks trained with cross entropy (CE) loss tend to find the most discriminative parts; (B) Part-based methods first locate several discriminative local parts and then extract features from them for classification; (C) Adversarial erasing methods encourage the model to learn more discriminative parts by progressively erasing the learned parts; (D) With assistant of Orthogonal loss, the student model implicitly mines the complementary and meaningful regions with the guidance of the teacher model.}
\label{fig:photo1}
\end{figure*}

Recently, convolution neural networks (CNNs) have undergone unprecedented success in visual representation learning. For classification tasks, it takes minimizing the cross-entropy loss (CE-loss) between the class labels and the network predictions as the optimization objective, as shown in Figure \ref{fig:photo1}(A). However, CE-loss usually makes the network to focus on the most discriminative region \cite{ge2019weakly} and ignore the other less significant but complementary local parts, which is not sufficient for FGVC. For instance, given an image containing a Corgi, the network may merely concentrate on its round butt, and ignore the remaining parts such as its cute head and short legs which also provide effective information for classification, as shown in Figure~\ref{fig:corgi}.

To overcome these limitations, it is widely accepted that the essential solution is to mine discriminative information from various complementary local regions \cite{wang2018learning,chen2019destruction}. In the early work, people resort help of heavy supervision to detect multiple discriminative parts for classification. It requires not only the category labels of the image but also additional manual annotation such as object part bounding boxes which consumes lots of human labor \cite{berg2013poof,lei2016fast}. Besides, the 
part annotations are hard to obtain during the inference phase, which reduces its usefulness and slows down the development of the community \cite{zheng2017learning}. Recently, weakly supervised detection or attention techniques become feasible substitutes, since only the category labels are needed for both the training and inference stage \cite{zheng2017learning,yang2018learning,wang2018learning,ma2019fine,du2020fine}. These methods can be roughly divided into two types: (i) part-based methods \cite{zheng2017learning,yang2018learning,wang2018learning,ma2019fine,chen2019destruction} which first locate several discriminative local parts and then extract features from them for classification, as shown in Figure \ref{fig:photo1} (B), and (ii) adversarial erasing methods \cite{DBLP:journals/corr/abs-1708-04552,Choe_2019_CVPR,8237643,Zhang_2018_CVPR} that encourage the model to learn more discriminative parts by progressively erasing the learned parts, as shown in Figure \ref{fig:photo1} (C). However, part-based methods may bring noises from background since most models have pre-defined the number of parts and multiple disciminative parts may not consistently occur in each image, and adversarial erasing methods suffer the same problem when too many object parts are erased.

In this paper, instead of explicitly locating the discriminative parts, we tend to address the limitation of CE-loss by knowledge transfer, where the student model implicitly mines the complementary discriminative regions with the guidance of the teacher models and then their results with complementary knowledge are combined together for the final recognition result, as shown in Figure \ref{fig:photo1}(D). Specifically, spatial attention maps are generated for both networks to represent their spatial concentrations where high response regions mean the discriminative parts and the low response regions indicate background. During the knowledge transfer stage, a orthogonal loss (OR-loss) is proposed to force the student model to mine discriminative information from the regions which are neither high response nor background on the spatial attention maps of the teacher models. Thanks to knowledge transfer, the student model not only obtains diverse and useful knowledge but also reduces the noises from background.

Main contributions of this paper can be summarized as follows:

\begin{enumerate}
    \item We apply knowledge transfer mechanism in FGVC, where several networks are trained in sequence, and all previously trained ones will serve as teacher models for the current student model. In this way, instead of focusing on the most discriminative region, the current model is forced to mine meaningful information from other regions and offer complementary knowledge.

    \item We propose a orthogonal loss (OR-loss), which can be used during knowledge transfer to guide the student network to mine discriminative information from the regions which neither have been focused by the teacher networks nor are background.
    
    \item Our method achieves SOTA performance on three widely used FGVC datasets, which demonstrates the effectiveness of our approach.
\end{enumerate}

\section{Related Work}

\subsection{Fine-grained Visual Classification} Due to large intra-class variations and small inter-class variations, FGVC is much more challenging than general classification \cite{fu2017look,zhang2019learning}, and the most effective solution for FGVC is to discover discriminative regions as many as possible. Recent works tackle FGVC from this perspective can be roughly divided into two types: (i) part-based methods \cite{zheng2017learning,yang2018learning,wang2018learning,ma2019fine,chen2019destruction}, and (ii) adversarial erasing methods \cite{Choe_2019_CVPR,8237643,Zhang_2018_CVPR}.

For part-based methods, the network first locates discriminative parts and then extracts features from these parts respectively. Fu et al. \cite{fu2017look} find region detection and fine-grained feature learning can reinforce each other, and build a series of networks which find discriminative regions for the next one. With similar motivation, Zhang et al. \cite{zhang2019learning} train several networks focusing on features of different granularities to produce diverse prediction distribution. Zheng et al. \cite{zheng2017learning} jointly learn part proposals and feature representations on each part. Nevertheless, these methods often suffer the problem that the number of found parts is pre-defined and discriminative parts may not consistently occur in each image. 

For adversarial erasing methods, the algorithm progressively erasing learned discriminative parts to encourage the network to learn more. In this way, not only the most discriminative region but also other informative parts are highlighted. Choe et al. \cite{Choe_2019_CVPR} utilize attention mechanism to drop the regions with high response. Lee et al. \cite{8237643} randomly hide patches to discover more object regions. Zhang et al. \cite{Zhang_2018_CVPR} find the complementary regions by two adversarial classifiers. However, it is hard to decide when the training process be stopped, and if too many object parts are erased the network tends to fix the background noise. 

In this paper, we propose a knowledge transfer based method and change the point of view by utilizing spatial information of both discriminative regions and background to encourage the model mining complementary information in a more robust way.

\subsection{Knowledge Distillation} Knowledge distillation Learning has been widely studied and applied in machine learning. It is a technology to distill the knowledge of one model to another. Hinton et al. \cite{hinton2015distilling} firstly propose the concept of dark knowledge based on teacher-student network, where teacher network is often a more complex network with desirable performance and fine generalization. With teacher's help the simpler student model with less parameter also has similar performance with teacher. Remero et al. \cite{romero2015fitnets} convey knowledge by feature maps, guiding student to acquire the ability to extract features. Zaforuyko et al. \cite{zagoruyko2017paying} propose the method that the student model is trained with the guidance of the teacher's attention map.

All aforementioned methods aim to make the student model has similar performance and fine generalization to the teacher model. However, in this paper, we guide the student model learn complementary knowledge with the teacher. We replace the KL divergence with the proposed OR-Loss and take spatial attention map as knowledge carrier. 

\section{Proposed Method}

In this section, we first revisit the CNN pipeline for general classification. Then two main parts of our knowledge transfer framework are introduced: the spatial attention maps which act as knowledge carrier, the orthogonal loss which induces the training of the student network.

\subsection{Revisiting CNN for classification}
Given an input image, the network first extracts features $F\in R^{C\times H\times W}$ which consist of $C$ channels with $H\times W$ spatial dimensions, and then feed into a classifier.

During training phase, minimizing the CE-loss between the labels and the predictions is employed as the optimizing objective. It has been proved that, with CE-loss only, the network is encouraged to locate the most discriminative region in the input image \cite{ge2019weakly} and ignores others. This natural character hinders the performance of CNN in FGVC tasks where various discriminative parts are sufficient. To address this limitation, in this paper, we train multiple networks in sequence with different concentrations that provide complementary knowledge during inference. Furthermore, network trained arbitrarily often extracts the most significant region. Hence, we propose a knowledge transfer framework with a newly proposed OR-Loss forcing the student model to mine complementary and meaningful information with the guidance of the teacher models.
\subsection{Spatial Attention Map}

During the knowledge transfer stage, our aim is transferring spatial information between two models, which makes spatial attention map an appropriate choice to be the information carrier since its response indicates spatial information of the model. The spatial attention map is obtained by the following procedure. First, the channel importance of a feature map is aggregated by global average pooling (GAP) operation. Then, we broadcast the GAP outputs along the channel dimension and element-wisely multiply them with the feature maps. At last, a channel-wise average pooling (CAP) operation is applied to the feature maps to compress the $3$D features into a $2$D spatial attention map $M$. In short, the process is denoted as follows:
\begin{align}
M=CAP(GAP(F) \otimes F),
\end{align}
where $\otimes$ denotes element-wise multiplication, and $M\in R^{1\times W\times H}$ represents the spatial attention map. 

Further more, the spatial attention map is normalized to standard normal distribution with zero mean. It is operated as follows:
\begin{align}
M_{norm}=\frac{M-M_{mean}}{M_{std}},
\end{align}
where $M_{mean}$ and $M_{std}$ are mean value and standard deviation value of the spatial attention map $M$, respectively. 

\begin{figure}[!t]
\centering
\includegraphics[height=4cm]{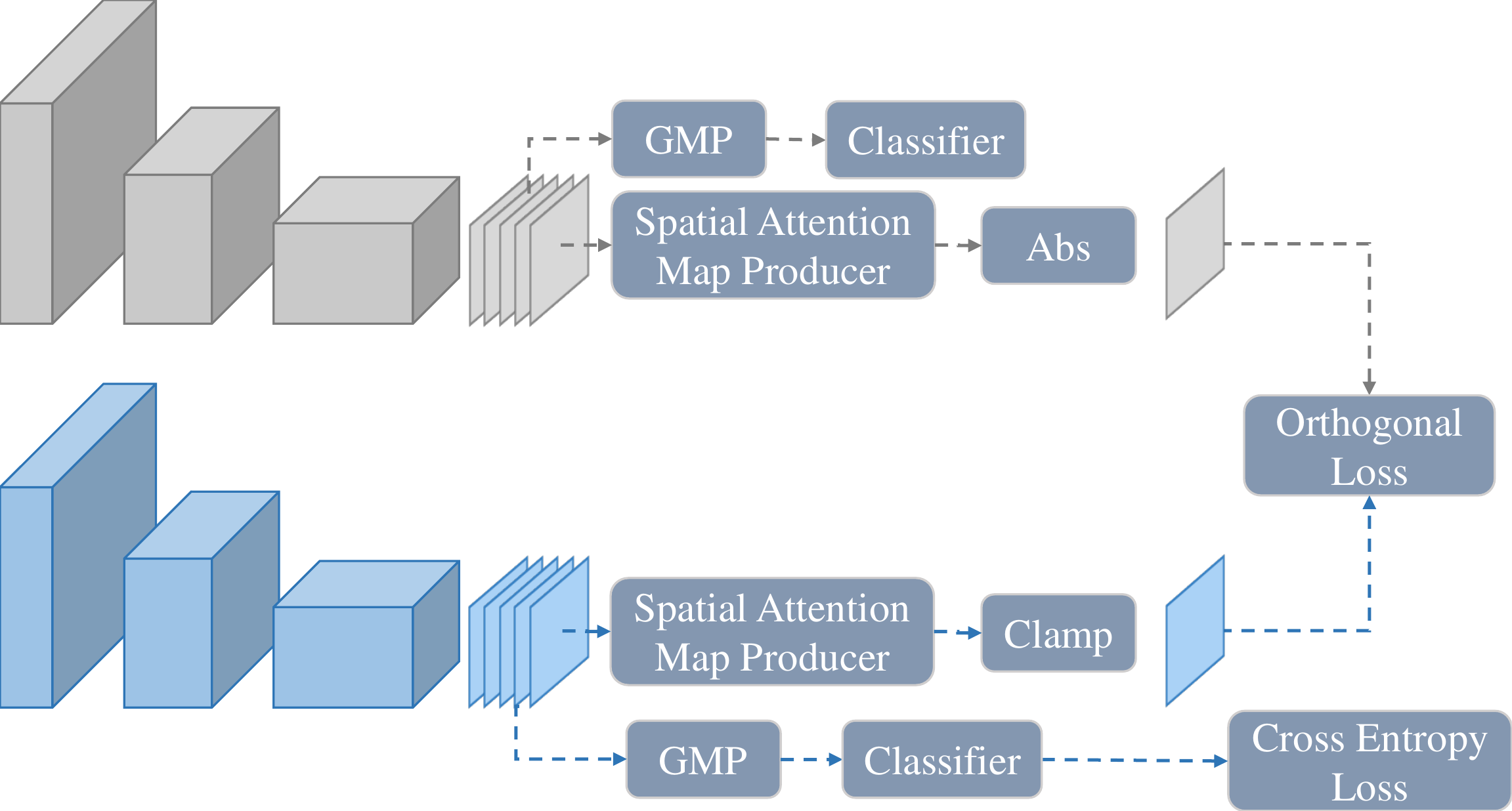}
\caption{The framework of our method. The upper components and the bottom components represent the forward propagation processes of the teacher model and the student model, respectively.}
\label{fig:photo2}
\end{figure}

\begin{figure}[!t]
\centering
\includegraphics[height=2.5cm]{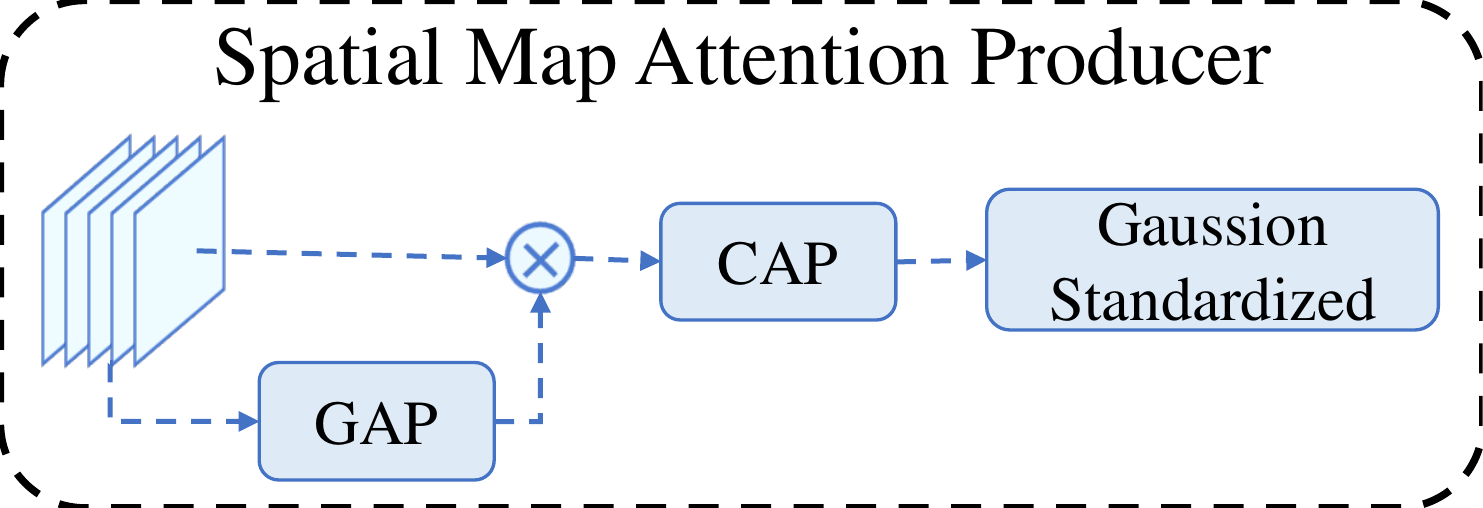}
\caption{The architecture of the spatial attention map producer.}
\label{fig:photo3}
\end{figure}

\subsection{Orthogonal loss} 
On the attention map, high response regions stand for the discovered discriminative parts and low response regions indicate background. To ensure the different concentrations between the teacher models and the student model, an orthogonal loss (OR-loss) is applied to their spatial attention maps, which guarantee them to be diverse. 

Furthermore, we hope the knowledge of background is also transferred to the student model to avoid the background noise being mined by mistake. Hence, the absolute value of the teacher attention map is used when computing the OR-loss:
\begin{align}
M_{teacher}^n=\left |M_{norm}  \right |.
\end{align}

Besides, only the high response regions of the student attention map need guidance of the OR-loss, so we only take the positive parts of the student attention map:
\begin{align}
M_{student}=max(M_{norm}, 0),
\end{align}
then the OR-loss can be denoted as:
\begin{align}
L_{OR}^{n}=M_{teacher}^n\otimes M_{student},
\end{align}
where $\otimes$ denotes element-wise multiplication. When the $N_{th}$ model is trained, where $N \geq 2$, all previous trained networks will serve as the teacher model, and $M_{teacher}^n$ denotes the knowledge carrier from the $n_{th}$ model network with $n\in\{1,N-1\}$. During knowledge transfer phase, CE-loss is also applied to encourage the model learning discriminative information under the orthogonal loss. A hyper-parameter $\alpha$ is introduced to balance the value of CE-loss and OR-loss, and the total loss $L_{total}$ is expressed as follows:

\begin{align}
L_{total}=L_{CE}+ \frac{1}{N-1} \sum_{n=1}^{N-1} \alpha L_{OR}^n,
\end{align}
where $N$ represents the ensemble system consists of $N$ models.

All teacher and student networks constituted the ensemble system. To make a comprehension use of each model's discriminative information, we average inference results and get the final classification result.

\section{Experiments}


\subsection{Datasets and Evaluation Protocol}
The proposed model is evaluated on three public FGVC benchmarks: Caltech-UCSD Birds (CUB) \cite{WahCUB_200_2011}, Stanford Cars (CAR) \cite{krause20133d}, and FGVC-Aircraft (AIR) \cite{DBLP:journals/corr/MajiRKBV13}. All our experiments are conducted without any bounding box annotations. To evaluate the performance of our method, we employ the top-1 accuracy as evaluation metric.


\subsection{Implementation Details} 

In this paper, all our experiments are performed on $4$ NVIDIA Geforce $1080$Ti GPUs with $12$G memory. Images are resized to $448\times 448$ before fed into the networks. Then we adopt random crop and random flip as the data augmentations. After that, we feed these images into ResNet$50$ or VGG$16$, initialized by the ImageNet \cite{russakovsky2015imagenet}. We use a mini-batch size of $32$. The network models are trained for $100$ epochs, using Stochastic Gradient Descent as optimizer and Batch Normalization \cite{ioffe2015batch} as regularizer. We initialize the learning rate of pretrained feature extractor and classifier to $0.001$ and $0.01$, respectively. And we use cosine annealing algorithm \cite{loshchilov2016sgdr} to gradually adjust learning rate every epoch. We empirically set momentum to $0.9$, weight decay to $5e^{-4}$, and the number of models $N$ to $5$.
\subsection{Comparisons with SOTA methods}

To verify the effectiveness of the proposed method, it is compared with other SOTA methods. Table \ref{tab:sota} illustrates the experiment results on CUB, CAR, and AIR datasets, respectively.

\begin{table}[!t]
  \centering
  \caption{Classification accuracy on CUB, CAR, and AIR. The best results on each dataset are in \textcolor{red}{red}, and the second best results are in \textcolor{blue}{blue}.}
  \vspace{+0.3cm}
    \begin{tabular}{|c|c|c|c|c|}
    \hline
 Method  & Backbone & \multicolumn{1}{l|}{CUB} & \multicolumn{1}{l|}{CAR} & \multicolumn{1}{l|}{AIR}\\
 \hline
 \hline
FT VGGNet \cite{wang2018learning} & VGG$19$ & $77.8$ & $84.9$ & $84.8$ \\
FT ResNet \cite{wang2018learning} & ResNet$50$ & $84.1$ & $91.7$ & $88.5$ \\
KA \cite{cai2017higher} & VGG$16$ & $85.3$ & $91.7$ & $88.3$ \\
KP \cite{cui2017kernel} & VGG$16$ & $86.2$ & $92.4$ & $86.9$ \\
MA-CNN \cite{zheng2017learning} & VGG$19$ & $86.5$ & $92.8$ & $89.9$ \\
RA-CNN \cite{fu2017look} & VGG$19$ & $85.3$ & 92.5 & - \\
DFL-CNN \cite{wang2018learning} & ResNet$50$ & $87.4$ & $93.1$ & $91.7$ \\
NTS-Net \cite{yang2018learning} & ResNet$50$ & $87.5$ & \textcolor{blue}{$93.9$} & $91.4$ \\
TASN \cite{8953519} & ResNet$50$ &  \textcolor{blue}{$87.9$} & $93.8$ & - \\
MC-Loss \cite{chang2019the} & ResNet$50$ & $87.3$ & $93.7$ & \textcolor{blue}{$92.6$} \\
PMA \cite{9103943} & ResNet$50$ & $87.5$ & $93.1$ & $90.8$ \\
   \hline
   \hline
Ours & VGG$16$ & $86.2$ & $91.9$ & $90.5$ \\
Ours & ResNet$50$ &  \textcolor{red}{$89.1$} & \textcolor{red}{$94.1$} & \textcolor{red}{$93.3$} \\
   \hline
    \end{tabular}%
  \label{tab:sota}%
\end{table}%

Our method achieves state-of-the-art (SOTA) performances on three datasets with ResNet$50$ as the backbone network. And when we use VGG$16$ as the base model, the proposed method also outperforms the baseline with a large margin of $8.4$\%, $7$\%, and $5.7$\%, which demonstrates the superiority of our framework can be generalized to any network architecture. As for representative part-based arts, MA-CNN \cite{zheng2017learning} adopts weakly-supervised object detection to locate different spatial regions for further prediction and RA-CNN \cite{fu2017look} creatively locates regions of different scales to obtain complementary information, which are suppressed by our framework on all of three datasets. For PMA \cite{9103943} supervised with only category label, which is a standard adversarial erasing framework for FGVC, we suppress it with significant improvements of $1.6\%$, $1.0\%$, and $2.5\%$ on all three datasets, which demonstrates the effectiveness of the proposed method. 

\subsection{Ablation Studies}

To select the optimal hyper-parameter $\alpha$ and number of models $N$, we conduct a series of experiments on the CUB dataset with ResNet$50$ as the base model, and report the results in Table \ref{tab:omega} and Table \ref{tab:n}, respectively.

\subsubsection{Effects of the hyper-parameter $\alpha$} 
It can be seen in Table \ref{tab:omega}, when the hyper-parameter $\alpha$ is set to $0$, the performance of ensemble system is not good enough. It indicates that with CE-loss, different networks only extract similar regions and ignore less informative parts which also benefit our final result. However, when the hyper-parameter $\alpha$ is between $0.05$ and $1$, the ensemble system obtains significant improvement, which verifies the validity of our method. The best result is obtained when $\alpha$ is set to $0.5$. Besides, when setting $\alpha$ to $5$, the performance drops a lot, since OR-loss demonstrates the training and hurts the optimization of CE-loss. Eventually, we set $0.5$ as the hyper-parameter $\alpha$.
\begin{table}[!t]
  \centering
        \caption{The performances of the proposed method with different loss weight $\alpha$. The best result is in \textcolor{red}{red}.}
    \vspace{+0.3cm}
    \begin{tabular}{|c|c|c|c|c|c|c|}
    \hline
   $\alpha$ & $0$ & $0.05$ & $0.1$ & $0.5$ & $1$ & $5$ \\
   \hline
   \hline
   Acc. & $87.2$ & $87.7$ & $87.9$ & \textcolor{red}{$88.5$} & $88.2$ & $87.2$ \\
   \hline
    \end{tabular}%
  \label{tab:omega}%
\end{table}%

\begin{table}[ht]
  \centering
   \caption{
   Comparison of our framework and the baseline ensemble system where all models are trained arbitrarily. Both the single model accuracy and the ensemble accuracy with different hyper-parameters $N$ are listed. The best results are in \textcolor{red}{red}.
   }
  \vspace{+0.3cm}
  \begin{adjustbox}{width=0.9\linewidth,center}
    \begin{tabular}{|c|c|c|c|c|c|c|}
    \hline
$N$ & $1$ & $2$ & $3$ & $4$ & $5$ & $6$ \\
 \hline 
\hline
Single (Base) & \textcolor{red}{$87.2$} & $86.8$ & $87.1$ & $87.0$ & $86.6$ & $87.0$ \\
\hline
Ensemble (Base) & $87.2$ & $87.8$ & $88.2$ & $88.3$ & \textcolor{red}{$88.6$} & $88.5$ \\
 \hline
  \hline
Single (Ours) & \textcolor{red}{$87.2$} & $86.1$ & $87.0$ & $86.5$ & $86.4$ & $86.6$ \\
 \hline
Ensemble (Ours) & $87.2$ & $88.5$ & $88.6$ & $88.8$ & \textcolor{red}{$89.1$} & \textcolor{red}{$89.1$}  \\
\hline
    \end{tabular}%
    \end{adjustbox}
  \label{tab:n}%
\end{table}%

\subsubsection{Effects of the Number of Models $N$}

As for the performance of each model, Table \ref{tab:n} shows the first model achieves the best performance in our framework, since it focuses on the most discriminative region of the input. Although the discriminative region is suppressed during the training phase, the accuracy of each student model does not significantly drop compared to baseline models trained arbitrarily, which illustrates that other region has discriminative information too.  

We conduct experiments about the number of models $N$ ranging from $1$ to $6$. As shown in Table \ref{tab:n}, when we increase the number of models from $1$ to $5$, the classification performance of our ensemble system receives consecutive gains and steadily exceeds the performance of the baseline ensemble system which contains $N$ models trained arbitrarily. It confirms the achievement of our motivation that mines and fuses complementary knowledge of less discriminative regions, with the guidance of both OR-loss and CE-loss. However, when $N$ is set to $6$, the accuracy of the ensemble system does not show any further improvements, which indicates the discriminative regions have already been excavated when $N=5$. At last, we set the number of models $N$ to $5$. 





\subsection{Visualization}
\begin{figure}[!t]
\centering
\includegraphics[height=5cm]{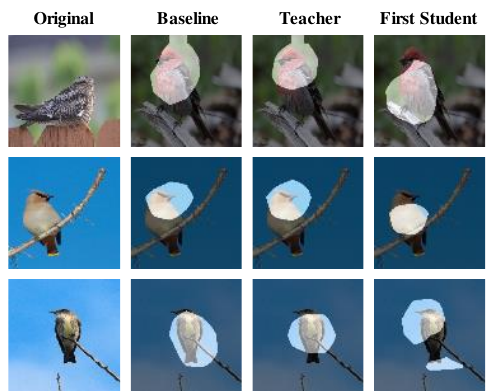}
\caption{Activation map of selected results on the CUB dataset on Resnet$50$. First column represents the original image. The following three columns show visualizations of baseline network, teacher network, and the first student network.}
\label{fig:photo4}
\end{figure}

Figure \ref{fig:photo4} displays original images and visualizations of the baseline network, the teacher network, and the first student network from Grad-CAM on CUB dataset based on ResNet$50$ model. Both of teacher and baseline network are trained with only CE-Loss. They only pay attention to the most discriminative region like bird's chest, and ignore lots of information of less discriminative regions. However, with assistance of the OR-loss, the student network mines complementary regions (e.g., bird's tail) from the teacher network. The visualization results indicate the effectiveness of OR-Loss of forcing the student network to focus on diverse regions.

\section{Conclusion}


In this work, we propose a knowledge transfer based approach for FGVC. Specifically, we train several models in sequence, and each of them are supervised by knowledge transferred from previous trained ones. During the training stage, under the guidance of both the CE-Loss and the proposed OR-Loss, the network is encouraged to find diverse and meaningful discriminative regions. Finally, to utilize the complementary knowledge of these models, we ensemble their inference outputs as the final prediction result. We demonstrate the superiority of our method and achieve the SOTA performance on CUB, CAR, and 
AIR datasets.
\bibliographystyle{IEEEbib}
\bibliography{icme2020template}

\end{document}